\documentclass[conference]{IEEEtran}
\IEEEoverridecommandlockouts
\usepackage{cite}
\usepackage{amsmath,amssymb,amsfonts}
\usepackage{algorithmic}
\usepackage{graphicx}
\usepackage{textcomp}
\usepackage{xcolor}
\usepackage{algorithm}
\usepackage{url}
\usepackage{booktabs}
\usepackage{array}
\usepackage{tabularx}
\usepackage{enumitem}
\usepackage{eso-pic}
\usepackage{flafter}
\begin{document}

\title{GSETS: Gaussian Smoothed Efficient Transition Sampling for Data-Driven Digital Twin Models}

\author{\IEEEauthorblockN{Ahmad Al-Khateeb\textsuperscript{1, 2}, Naheel Kamal\textsuperscript{3}, Eman Hammad\textsuperscript{1, 2}, Sertac Bayhan\textsuperscript{3}, Haitham Abu-Rub\textsuperscript{3,4} }
\textit{\textsuperscript{1}iSTAR Lab, Texas A\&M University, College Station, TX, USA}\\
\textit{\textsuperscript{2}Dept. of Electrical \& Computer Engineering, Texas A\&M University, College Station, TX, USA}\\
\textit{\textsuperscript{3}Qatar Environment and Energy Research Institute, Hamad Bin Khalifa University, Doha, Qatar}\\
\textit{\textsuperscript{4}Dept. of Electrical \& Computer Engineering, Hamad Bin Khalifa University, Doha, Qatar
}\\
Corresponding author: ahmad.alkhateeb@tamu.edu
}

\maketitle
\AddToShipoutPictureBG*{\AtPageLowerLeft{\put(0,22){\makebox[\paperwidth]{\parbox{0.84\paperwidth}{\centering\scriptsize
\copyright~2026 IEEE. Personal use of this material is permitted. Permission from IEEE must be obtained for all other uses, in any current or future media, including reprinting/republishing this material for advertising or promotional purposes, creating new collective works, for resale or redistribution to servers or lists, or reuse of any copyrighted component of this work in other works.}}}}}

\begin{abstract}
Data-driven models and digital twins are fundamentally limited by the quality of their training data, and acquisition cost often constrains how much can be collected. Conventional one-shot methods such as uniform grid and random sampling determine where samples are placed in the operating space, but not the order in which they are visited, leaving the step transitions that dictate transient behavior unoptimized. Adaptive sampling partially addresses this, but only through a closed loop that retrains the surrogate at every step, which is impractical when data acquisition and training are structurally decoupled. This paper introduces Gaussian-Smoothed Efficient Transition Sampling (GSETS), a one-shot trajectory generation framework that maximizes transition diversity and information density per sample. Base GSETS performs a memory-augmented random walk over a discretized state space, tracking prior exploration in a global visitation heatmap updated by Gaussian penalties and selecting successive targets via a temperature-controlled Softmax. Hierarchical GSETS extends this from node-level to edge-level penalization using a 4D transition tensor and a dual-kernel update, steering neighboring states away from paths parallel to recently explored vectors. The framework is evaluated by training a DNN surrogate with multi-scale time-window features on parallel-connected, dual power-sharing T-type rectifiers. Hierarchical GSETS reduces DC-link voltage tracking RMSE by 21\% relative to random sampling under an identical budget, and 350 GSETS samples outperform grid and random baselines allocated nearly three times the data (N = 1000).
\end{abstract}

\begin{IEEEkeywords}
Digital Twins, Surrogate Modeling, Power Electronics, Neural Networks, Sampling Efficiency
\end{IEEEkeywords}

\section{Introduction}
The acquisition of training data is one of the most common and impactful problems faced when training data-driven models. It is the foundation that the entire model depends on, as all data-driven models are inherently limited by the quality of information in their data. Thus, the concern of overfitting or underfitting the model is always present when training, especially for deep learning models, which are known to be more susceptible to overfitting~\cite{li2024keeping}. While these challenges with generalization have been acknowledged and extensively studied~\cite{li2024keeping}, traditional regularizations are often supplemented by dynamic data collection frameworks. Specifically, adaptive sampling has emerged as a prominent solution, where new training data is iteratively generated based on the current model performance metrics~\cite{long2021multi,lagou2023adaptive}. The objective is to maximize surrogate model accuracy while minimizing the computational and time-intensive costs of simulating additional data points.

However, adaptive sampling introduces a computational bottleneck, requiring a lengthy, closed-loop pipeline where the underlying model must be repeatedly trained and re-evaluated~\cite{li2024keeping}. Such an approach becomes impractical when the data acquisition phase and the training phase are structurally decoupled, making closed-loop integration highly complex or cost-prohibitive. Conversely, one-shot sampling techniques offer an appealing alternative by completely isolating data collection from the training loop, thereby requiring only a single, offline data acquisition phase. Despite their operational simplicity, these one-shot methods are often overlooked due to poor sample efficiency, demanding excessive data volumes to adequately explore a sample space~\cite{long2021multi}. 

While considerable attention is dedicated to tuning neural network hyperparameters and architectures to maximize performance out of static datasets~\cite{iranfar2021multiagent,liu2021survey}, comparatively little emphasis is placed on optimizing the trajectory generation logic that captures the data in the first place, with uniform grid discretization and random sampling remaining the dominant strategies in power electronics and power systems surrogate modeling~\cite{bayindir2024design, xie2023transparent,
wang2025real}.

This trajectory-based approach is particularly relevant when there is direct control over how data is sampled and generated, making it highly advantageous when either data collection or model training is expensive. Under these constraints, the value of high-quality, dense, and diverse data becomes readily apparent. This makes the framework directly applicable to digital twin (DT) development, where developers typically possess full control over the physical system or its high-fidelity simulation environment to generate target training sequences~\cite{bayindir2024design, sharida2023digital}. While contemporary power electronic surrogate models often rely on uniform datasets~\cite{bayindir2024design, xie2023transparent} or Latin Hypercube Sampling~\cite{wang2025real} to map operational spaces, these standard techniques treat data points as independent, isolated states. Although this static assumption suffices for steady-state mapping, it fails for non-linear, temporal systems where the transition step-changes directly dictate transient stability and dynamic system behavior.

This methodology shares conceptual parallels with artificial potential field methods~\cite{luo2022reliable}, reinforcement learning, and continuous trajectory sampling frameworks~\cite{lin2021continuous}. However, while there are commonalities between efficient transition sampling and traditional spatiotemporal sampling, key differences redefine the core objective. In typical spatial applications, like robotic field mapping, data collection is constrained by physical sequential trajectories where robots cannot simply jump across the domain, forcing sample selection to incur penalties based on geometric distance or localized obstacle and collision buffers~\cite{luo2022reliable, newaz2023decentralized}. Conversely, when sampling transitions to capture dynamic profiles in an abstracted control parameter space, there is a strong incentive to encourage a wide variety of spatial distances between consecutive target states. In the proposed framework, large spatial jumps between transitions incur no mechanical penalty or physical travel limitations, allowing the system to intentionally select drastic step-changes to thoroughly map the complete transient behavior of the system.

Therefore, a gap remains across both paradigms. Grid, random, and adaptive methods all decide where points are placed, but none decide the order in which they are visited. For non-linear temporal systems, where step-changes dictate transient stability and dynamic response, this leaves the most informative part of the data unoptimized. Adaptive sampling cannot readily fill this role either, since selecting transitions in a closed loop requires retraining the surrogate at every step. What is needed is an offline method that targets transition diversity directly.

This work reformulates one-shot trajectory generation as a curiosity-driven exploration problem. The framework retains an explicit memory of prior exploration and biases each successive target toward under-sampled regions and, in its hierarchical form, under-sampled directions of movement. It is validated on a surrogate for a dual power-sharing T-type rectifier, whose power-sharing adjustments produce highly non-linear transients that are expensive to simulate continuously. The main contributions are:

\begin{enumerate}[leftmargin=*,itemsep=1pt,topsep=3pt,parsep=0pt]
    \item Base GSETS, a memory-augmented random walk using a global Gaussian visitation heatmap with temperature-controlled Softmax selection.
    \item Hierarchical GSETS, extending node-level to edge-level penalization via a 4D transition tensor and a dual-kernel update whose origin-side
    spillover steers neighboring states away from paths parallel to recently explored vectors.
    \item A multi-scale feature scheme using elapsed-time sliding windows rather than fixed sample lags, condensing high-frequency waveforms into 36 features and keeping the surrogate agnostic to hardware sampling rates.
    \item An evaluation showing a 21\% reduction in $V_{dc}$ tracking RMSE at an identical budget, with 350 GSETS samples outperforming grid, shuffled-grid, and random baselines given nearly three times the data.
\end{enumerate}

\section{System Model}
\begin{figure}
    \centering
    \includegraphics[width=\linewidth]{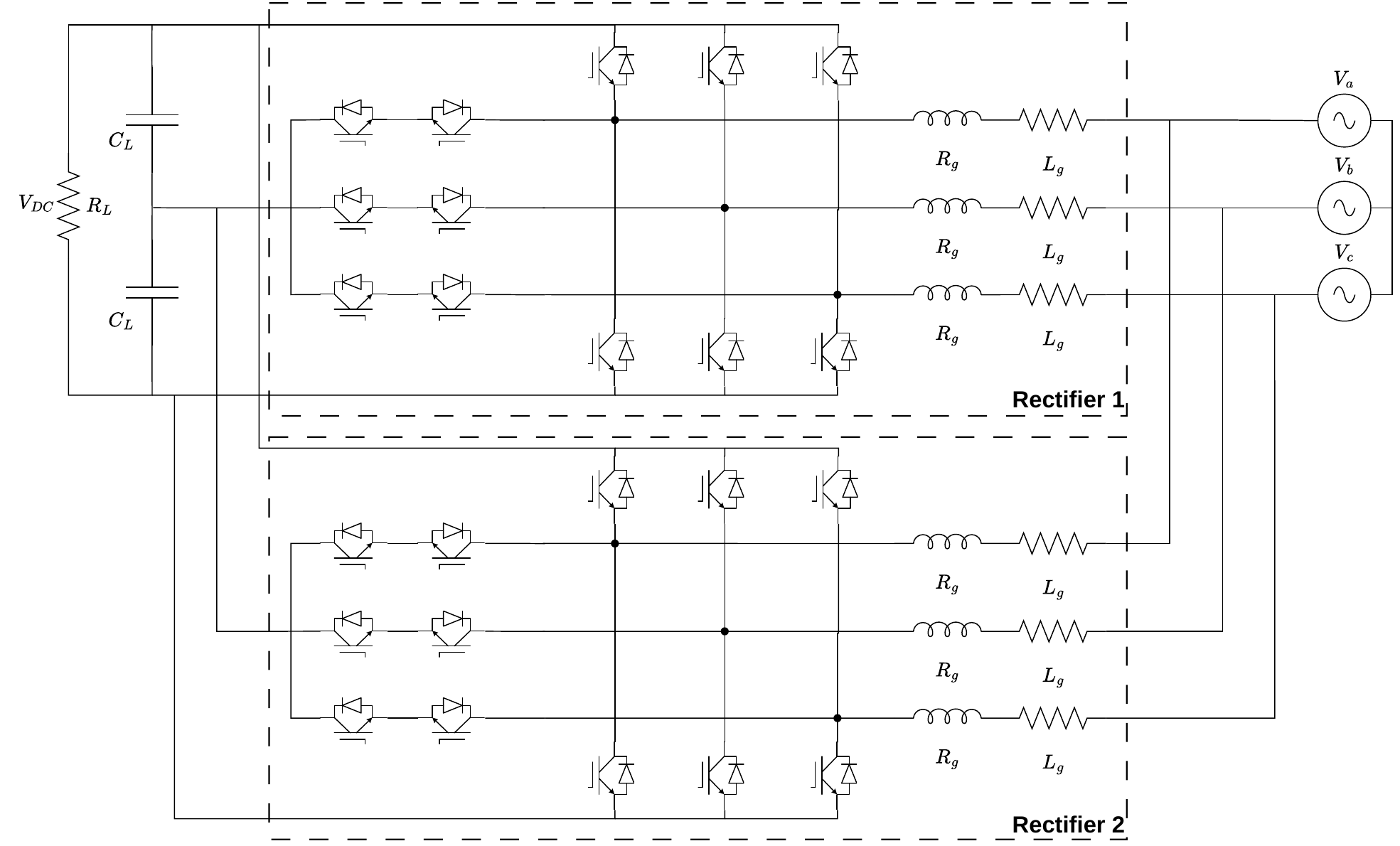}
    \caption{Power sharing T-type rectifiers system model.}
    \label{fig:system_model}
\end{figure}

The modeled system is a dual power-sharing rectifier setup, as illustrated in Fig.~\ref{fig:system_model}. It consists of two T-type rectifiers connected in parallel to support a common load. This topology is favored for its modularity and inherent scalability, making it highly suitable for high-power applications such as Electric Vehicle (EV) charging infrastructure~\cite{long2023impartial}. The system was first modeled in Simulink with Inverse Model Predictive Control (IMPC), to serve as the ground truth for data generation. It should be noted that the proposed sampling methodology is independent of the underlying modeling medium; it should maintain its performance regardless of whether the ground truth is a software simulation or a real-world hardware environment, as in~\cite{bayindir2024design,sharida2023digital}. 

\begin{table}[t]
    \centering
    \caption{Experimental Setup Parameters}
    \begin{tabular}{cccc} \hline
        Symbol & Value & Symbol & Value\\ \hline
        $V_{abc}$ & 100-120 Vrms & $V_{DC}$ & 400 V\\
        $P_{Load}$ & 1-30 kW & $T_s$ & 10 $\mu$s \\ \hline
    \end{tabular}
    \label{tab:systemParam}
\end{table}

The primary operational constraints and electrical parameters are listed in Table~\ref{tab:systemParam}. The system's operational state is dictated by two independent physical variables that form the continuous state space $\mathcal{S}$: total requested load power ($P_{Load}$) and the power-sharing ratio ($PSR$). The $PSR$ can be dynamically adjusted throughout operation to adjust the portion of the load handled by each converter, where:
\begin{equation}
    P_{\text{rect1}} = PSR \cdot P_{\text{Load}}, \quad P_{\text{rect2}} = (1 - PSR) \cdot P_{\text{Load}}
\end{equation}
A $PSR$ of 0.5 denotes the load being shared equally by both rectifiers, whereas boundary ratios of 0 or 1 force a single rectifier to sustain the entire load demand. 

\section{Methodology}

\begin{figure*}[htbp]
    \centering
    \includegraphics[trim=0 .6cm 0 0, clip, width=\textwidth]{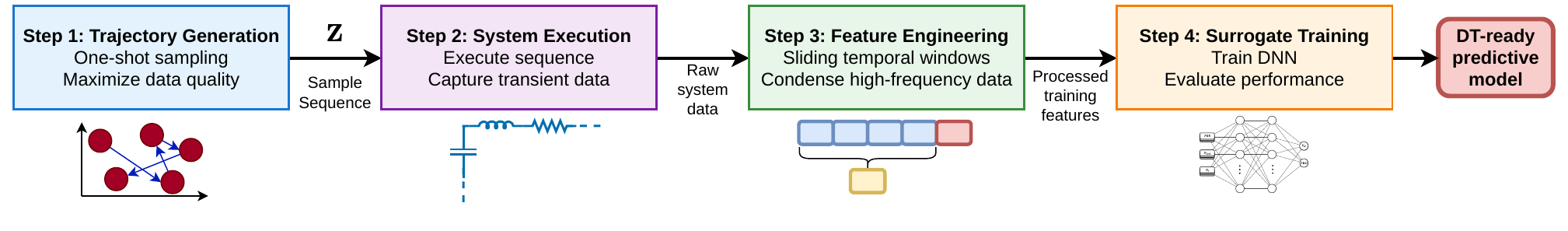}
    \caption{Proposed work pipeline}
    \label{fig:framework_pipeline}
\end{figure*}

The complete operational pipeline of the surrogate modeling framework utilized to demonstrate the proposed method is illustrated in Fig.~\ref{fig:framework_pipeline}. The architecture is modular and agnostic to the underlying hardware domain, separated into four sequential phases: (i) an offline, curiosity-driven trajectory planning phase that establishes an ordered step sequence $\mathbf{Z}$; (ii) a system execution phase where the planned sequence is applied to a target physical system or high-fidelity simulation model; (iii) a multi-scale feature engineering phase that condenses continuous time-series waveforms using sliding temporal windows; and (iv) a surrogate training phase where a machine learning model is trained on the previously generated data. The technical details of each pipeline component will be expanded on in this section, starting with the trajectory generation methods.

\subsection{Conventional Sampling Methods}
To identify the benefits of the proposed method, it is compared against two conventional data sampling methods: grid and random sampling. While these methods are naive from an algorithmic perspective, they are ubiquitous for data-driven modeling as they reliably produce diverse data without much effort.

\subsubsection{Uniform Grid Sampling}
The uniform grid baseline operates as a deterministic discretization of the continuous parameter boundaries. Given a two-dimensional state space $\mathcal{S}$ defined by the limits of the load power $P_{\text{Load}}$ and the power-sharing ratio $PSR$, the parameter ranges are evenly partitioned. While this strategy ensures uniform spatial distribution across the steady-state operating space, its rigid determinism completely omits any variation in step magnitude or direction. Therefore, the sequential transitions between adjacent grid coordinates remain highly repetitive, failing to capture a diverse variety of transient behaviors. To compensate for this, a second version where the sequence of the grid samples is shuffled is also included (denoted as Grid shf). 

\subsubsection{Independent Random Sampling}
Independent random sampling is a purely stochastic collection of non-sequential data. At each discrete trajectory sample interval $k$, the target operational variables are chosen independently from a uniform continuous distribution within the operating bounds:
\begin{equation}
    P_{\text{Load}, k} \sim \mathcal{U}(P_{\text{min}}, P_{\text{max}}), \quad PSR_k \sim \mathcal{U}(PSR_{\text{min}}, PSR_{\text{max}})
\end{equation}

\begin{figure}
    \centering
    \includegraphics[width=0.7\linewidth]{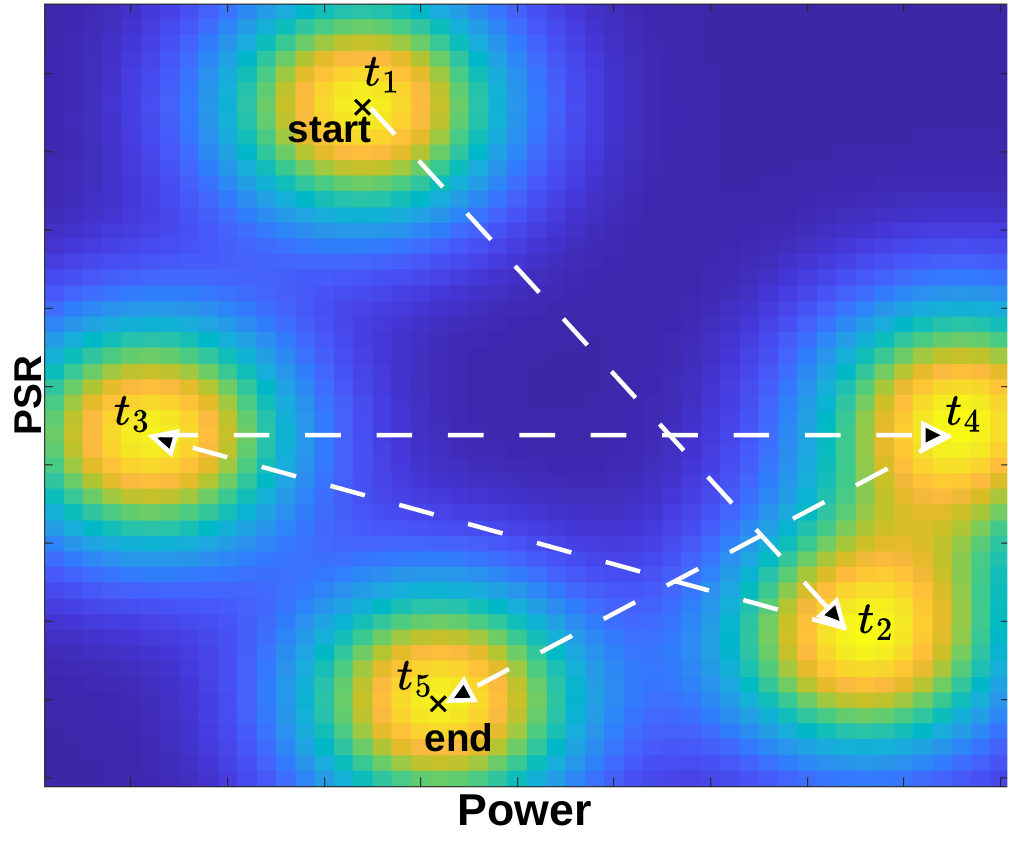}
    \caption{Gaussian heatmap visualization of 5 consecutive samples}
    \label{fig:heatmap}
\end{figure}

\subsection{Base GSETS}
The Base GSETS framework operates as a memory-augmented random walk over a discretized 2D state space grid $\mathcal{S} \subset \mathbb{R}^2$. To disincentivize the trajectory generator from repeatedly visiting the same regions, the algorithm tracks exploration using a global visitation heatmap $H_k(s)$. At step $k$, when the walker occupies state $s_k$, a Gaussian penalty is added to the heatmap:
\begin{equation}
    H_{k+1}(s) = H_k(s) + \exp\left(-\frac{\|s - s_k\|^2}{2\sigma_{\text{global}}^2}\right)
    \label{eq:base_heatmap}
\end{equation}
where $\sigma_{\text{global}}$ defines the spatial bandwidth of the penalty. This operation mathematically marks the region around state $s_k$ as ``visited'', increasing its resistance to future exploration, as visually mapped in Fig.~\ref{fig:heatmap}.

The next state is sampled using a Softmax function applied to the negated heatmap values:
\begin{equation}
    P(s_{k+1} = s) = \frac{\exp(-H_{k+1}(s) / \tau)}{\sum_{s' \in \mathcal{S}} \exp(-H_{k+1}(s') / \tau)}
    \label{eq:base_sample}
\end{equation}
The temperature parameter $\tau$ controls the strictness of this curiosity. As $\tau \to 0$, the selection probability forcefully pushes the walker to the absolute coldest (least visited) states, while higher values allow for a more random walk. The full procedure is outlined in Algorithm~\ref{alg:base_gsets}.

\begin{algorithm}[htbp]
\caption{Base GSETS Trajectory Generation}
\label{alg:base_gsets}
\begin{algorithmic}[1]
\REQUIRE State space grid $\mathcal{S}$, Trajectory length $N$, Curiosity parameter $\tau$, Global smoothing $\sigma_{\text{global}}$
\STATE Initialize global heatmap array $H \leftarrow \mathbf{0}_{|\mathcal{S}|}$
\STATE Initialize trajectory sequence array $\mathbf{Z} \leftarrow \mathbf{0}_N$
\STATE Sample initial state $s_0 \sim \mathcal{U}(\mathcal{S})$
\STATE $s_k \leftarrow s_0$
\FOR{$k = 1$ \TO $N$}
    \STATE $\mathbf{Z}[k] \leftarrow s_k$ 
    \FORALL{$s \in \mathcal{S}$}
        \STATE $G(s) \leftarrow \exp\left(-\|s - s_k\|^2 / 2\sigma_{\text{global}}^2\right)$
        \STATE $H(s) \leftarrow H(s) + G(s)$
    \ENDFOR
    \STATE Compute logits: $L \leftarrow -H / \tau$
    \STATE Compute transition probabilities $P \leftarrow \text{Softmax}(L)$ 
    \STATE Sample next state $s_{k+1} \sim P$
    \STATE $s_k \leftarrow s_{k+1}$
\ENDFOR
\RETURN $\mathbf{Z}$
\end{algorithmic}
\end{algorithm}

\subsection{Hierarchical GSETS}
While Base GSETS achieves good sample space coverage, node-level penalization does not account for the direction of movement. As a result, the walker may traverse similar paths, creating redundant trajectories that do not add much informational value. To fix this, Hierarchical GSETS tracks exploration using a 4D transition tensor $\mathcal{T} \in \mathbb{R}^{|\mathcal{S}| \times |\mathcal{S}|}$ to enforce edge-level penalization.

When selecting the next state from the current step $s_k$, the framework samples strictly from the 2D matrix row slice corresponding to that state:
\begin{equation}
    P(s_{k+1} = s \mid s_k) = \frac{\exp(-\mathcal{T}_k(s_k, s) / \tau)}{\sum_{s' \in \mathcal{S}} \exp(-\mathcal{T}_k(s_k, s') / \tau)}
    \label{eq:hier_sample}
\end{equation}
This ensures the walker explicitly considers paths previously taken from its exact current location, rather than just checking global node density.

To prevent the framework from generating parallel paths in continuous space, a Dual-Kernel Update rule is applied immediately after moving from origin step $s_k$ to destination step $s_{k+1}$:
\begin{equation}
\begin{aligned}
    \mathcal{T}_{k+1}(s_i, s_j) = &\,\mathcal{T}_k(s_i, s_j) \,+ \\
    &\exp\left(-\frac{\|s_i - s_k\|^2}{2\sigma_{\text{orig}}^2}\right) \exp\left(-\frac{\|s_j - s_{k+1}\|^2}{2\sigma_{\text{dest}}^2}\right)
\end{aligned}
\label{eq:hier_update}
\end{equation}
Here, the destination kernel ($\sigma_{\text{dest}}$) penalizes the arrival node, while the origin kernel ($\sigma_{\text{orig}}$) applies a spatial spillover to neighboring departure nodes. This mechanism teaches adjacent states to avoid paths running parallel to recently explored vectors. 

To maintain computational efficiency, a threshold parameter $\epsilon$ restricts the tensor updates strictly to localized neighbors where the origin kernel value is non-negligible, as shown in Algorithm~\ref{alg:hierarchical_gsets}.

\begin{algorithm}[htbp]
\caption{Hierarchical GSETS (Edge-Level Penalization)}
\label{alg:hierarchical_gsets}
\begin{algorithmic}[1]
\REQUIRE State space grid $\mathcal{S}$, Trajectory length $N$, Curiosity parameter $\tau$
\REQUIRE Kernel scales: $\sigma_{\text{dest}}$, $\sigma_{\text{orig}}$
\REQUIRE Computational threshold $\epsilon$
\STATE Initialize 4D transition tensor $\mathcal{T} \leftarrow \mathbf{0}_{|\mathcal{S}| \times |\mathcal{S}|}$
\STATE Initialize trajectory sequence array $\mathbf{Z} \leftarrow \mathbf{0}_N$
\STATE Sample initial state $s_0 \sim \mathcal{U}(\mathcal{S})$
\STATE $s_k \leftarrow s_0$
\FOR{$k = 1$ \TO $N$}
    \STATE $\mathbf{Z}[k] \leftarrow s_k$
    \STATE Extract local row matrix: $M \leftarrow \mathcal{T}(s_k, :)$
    \STATE Compute transition probabilities:\\ $P \leftarrow \text{Softmax}(-M / \tau)$
    \STATE Sample next state $s_{k+1} \sim P$
    
    \STATE Compute destination kernel vector $G_{\text{dest}} \in \mathbb{R}^{|\mathcal{S}|}$ around $s_{k+1}$
    \STATE Compute origin kernel vector $G_{\text{orig}} \in \mathbb{R}^{|\mathcal{S}|}$ around $s_k$
    
    \FORALL{$s_i \in \mathcal{S}$ \textbf{where} $G_{\text{orig}}(s_i) > \epsilon$}
        \STATE $W \leftarrow G_{\text{orig}}(s_i)$
        \FORALL{$s_j \in \mathcal{S}$}
            \STATE $\mathcal{T}(s_i, s_j) \leftarrow \mathcal{T}(s_i, s_j) + W \cdot G_{\text{dest}}(s_j)$
        \ENDFOR
    \ENDFOR
    \STATE $s_k \leftarrow s_{k+1}$
\ENDFOR
\RETURN $\mathbf{Z}$
\end{algorithmic}
\end{algorithm}

\subsection{Neural Network Model Design}
\begin{figure}
    \centering
    \includegraphics[width=0.9\linewidth]{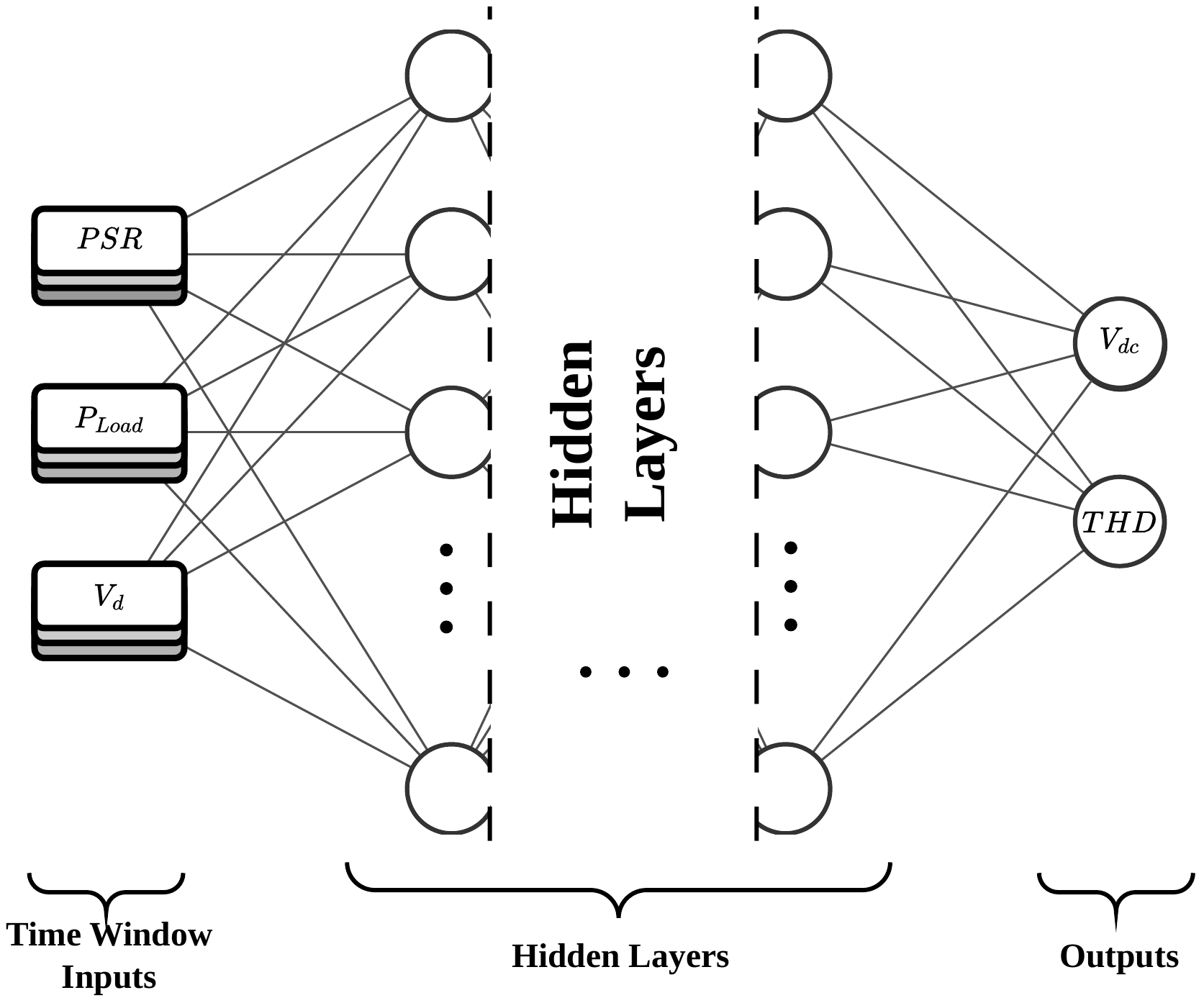}
    \caption{DNN with time window inputs}
    \label{fig:dnn_arch}
\end{figure}

To evaluate the performance of the different sampling methods a Deep Neural Network (DNN) surrogate model is developed. The DNN consists of 5 fully connected hidden layers: $[512, 256, 128, 64, 32]$ with dropout layers of rate $0.3$ in between~(Fig.~\ref{fig:dnn_arch}). The surrogate maps three measured signals ($PSR$, $P_{\text{Load}}$, and the d-axis grid voltage $V_d$) to two predicted outputs: the DC-link voltage $V_{dc}$ and the total harmonic distortion (THD) of the input currents.

As the power electronics simulation runs at a high sampling rate ($T_s = 10\,\mu\text{s}$), using standard sample-lagged inputs (tracking the last few discrete samples) is impractical. For instance, a window of a few dozen lagged samples captures only a fraction of a millisecond, which represents an incomplete picture regarding system transient dynamics.

\begin{figure}
    \centering
    \includegraphics[width=\linewidth]{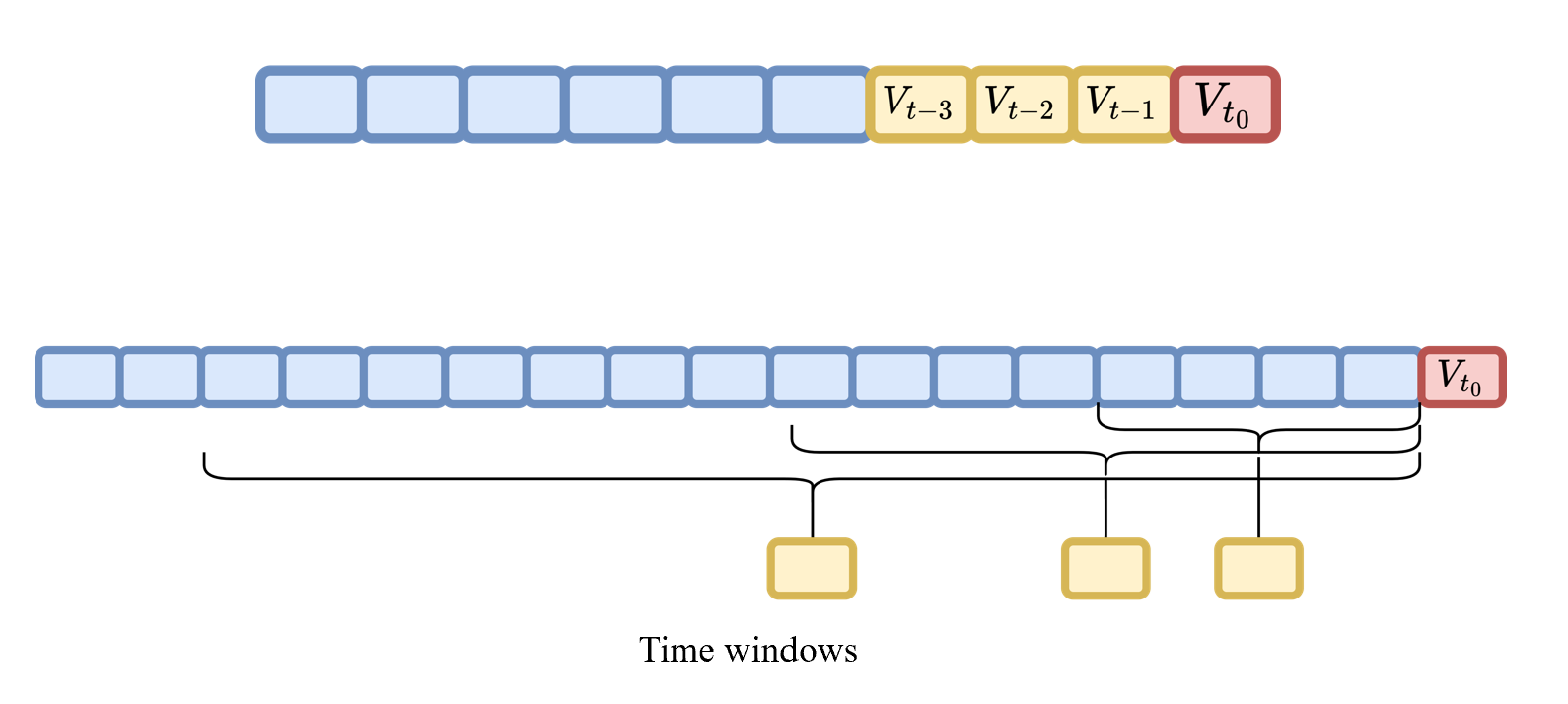}
    \caption{Time windows for features}
    \label{fig:time_windows}
\end{figure}

To resolve this, feature extraction is defined using sliding time windows over explicit elapsed time intervals rather than static sample counts (Fig.~\ref{fig:time_windows}). This approach condenses high-frequency data into a manageable feature set. It also keeps the model agnostic to physical sampling discrepancies, such as communication delays or variable hardware execution rates.

For an elapsed time window of duration $\tau$, the engineered feature vector is defined by a function $f(x_{t-\tau}, x_t)$ evaluating the waveform characteristics up to the current inference time $t$. Three time window functions were selected: (i) Mean, (ii) Standard Deviation, and (iii) Delta ($f_{\Delta}$).
The Delta function tracks the absolute rate of change across the window boundary, expressed as:
\begin{equation}
    f_{\Delta} = \frac{x_t - x_{t-\tau}}{\tau}
    \label{eq:delta}
\end{equation}

Four distinct time window intervals are chosen to capture multi-scale system behavior: $100\,\text{ms}$, $10\,\text{ms}$, $1\,\text{ms}$, and $0.1\,\text{ms}$. Applying these three operations for each of the three inputs across the selected intervals results in a total of 36 engineered input features, ensuring the DNN accurately maps both long-term settling trends and rapid localized switching transients.

\subsection{Experimental Setup and Parameter Configuration}
Data generation was executed in MATLAB/Simulink, holding system parameters constant for $0.25\,\text{s}$ per step to capture the complete transient profile. The continuous state space of $P_{\text{Load}}$ and $PSR$ is bound by a discrete $30 \times 30$ coordinate grid ($|\mathcal{S}| = 900$ states).

Each sampling method was evaluated across two training scales: a $1,000$-step training set paired with a $350$-step validation set, and a scaled-down $350$-step training set paired with a $90$-step validation set for comparative analysis. Validation sets consist of independent transitions used for hyperparameter tuning and early stopping. Model generalization is evaluated against a common test dataset comprising $2,000$ independent random transition steps.

Both GSETS variants utilize a curiosity temperature of $\tau = 1.0$ over the $900$-state tracking layout. Spatial boundaries are governed by explicit Gaussian kernel scales: Base GSETS applies a uniform node penalty scale of $\sigma_{\text{global}} = 2.0$, while Hierarchical GSETS decouples edge-level updates into a destination scale of $\sigma_{\text{dest}} = 2.0$ and an origin spillover scale of $\sigma_{\text{orig}} = 2.0$.

\section{Results and Discussion}
The performance comparison of the sampling algorithms across different dataset sizes ($N=350$ and $N=1000$) is summarized in Table~\ref{tab:performance_metrics}. The results reveal a clear performance improvement when shifting from conventional sampling methodologies to the trajectory-based GSETS framework. Standard deterministic grid sampling exhibits the highest error, yielding a poor $V_{dc}$ RMSE of 24.0697 at 350 samples. Shuffling the grid sequence (Grid shf) disrupts the rigid sequence bias and significantly lowers the error to 10.9174. This demonstrates the importance of the sequence of transitions, as small steps to adjacent nodes do not accurately represent the more significant transitions that can result from shuffling the order. However, both methods remain structurally inferior to independent random sampling (8.8837), likely because random operations are entirely unconstrained by a fixed spatial lattice. 
Integrating memory-augmented spatial exploration yields immediate performance improvement. The node-level curiosity framework (Base GSETS) consistently secures the second-best performance across nearly all metrics at both dataset scales. At $N=350$, the Base GSETS approach drops the $V_{dc}$ RMSE further to 7.5590 and the MAE to 3.6010, demonstrating that driving a trajectory toward under-explored spatial neighborhoods captures significantly higher information density than naive stochastic baselines.
In contrast, the proposed Hierarchical GSETS framework achieves the highest overall accuracy, minimizing the $V_{dc}$ RMSE to 7.0194 and MAE to 3.2122 at 350 samples. This represents a 21.0\% reduction in tracking error compared to conventional random sampling under an identical sample budget. This highlights the efficacy of edge-level tracking via the 4D transition tensor and dual-kernel update, which actively avoids redundant parallel trajectories. Furthermore, the results demonstrate remarkable sample efficiency. The Hierarchical GSETS model trained on a sparse budget of only 350 samples ($V_{dc}$ RMSE = 7.0194) directly outperforms standard Grid, Grid shf, and Random sampling even when those conventional methods are allocated nearly three times the data volume ($N = 1000$). When the GSETS budget is increased to match 1000 samples, the tracking error scales down further to 6.1213, confirming the capacity of the framework to consistently maximize information density per sample as the data allocation scales.

\begin{table}[htbp]
    \centering
    \caption{Performance Comparison Across Dataset Sizes (bold denotes the best performance; underline denotes the second best).}
    \label{tab:performance_metrics}
    \begin{tabularx}{\linewidth}{llXXXX}
        \toprule
        \textbf{Size} & \textbf{Method} & 
        \begin{tabular}[c]{@{}c@{}}\textbf{$V_{dc}$}\\ \textbf{RMSE}\end{tabular} & 
        \begin{tabular}[c]{@{}c@{}}\textbf{$V_{dc}$}\\ \textbf{MAE}\end{tabular} & 
        \begin{tabular}[c]{@{}c@{}}\textbf{THD}\\ \textbf{RMSE}\end{tabular} & 
        \begin{tabular}[c]{@{}c@{}}\textbf{THD}\\ \textbf{MAE}\end{tabular} \\ 
        \midrule
        350  & Grid     & 24.0697 & 7.9854 & 0.1438 & 0.0522 \\
             & Grid shf & 10.9174 & 4.7600 & 0.0719 & 0.0381 \\
             & Random   & 8.8837 & 3.9817 & 0.0530 & \underline{0.0275} \\
             & Base GSETS    & \underline{7.5590} & \underline{3.6010} & \underline{0.0458} & \underline{0.0275} \\
             & Hier. GSETS & \textbf{7.0194}  & \textbf{3.2122} & \textbf{0.0453} & \textbf{0.0243} \\
        \midrule
        1000 & Grid     & 23.3066 & 8.3170 & 0.1299 & 0.0460 \\
             & Grid shf & 8.4559  & 3.6768 & 0.0519 & 0.0289 \\
             & Random   & 7.1962 & 3.2844 & \underline{0.0459} & \textbf{0.0230} \\
             & Base GSETS     & \underline{6.6644}  &  \underline{3.2266}  &  0.0474  &  0.0286 \\
             & Hier. GSETS & \textbf{6.1213}  & \textbf{2.9762} & \textbf{0.0428} & \underline{0.0236} \\
        \bottomrule
    \end{tabularx}
\end{table}

\begin{figure}
    \centering
    \includegraphics[width=1\linewidth]{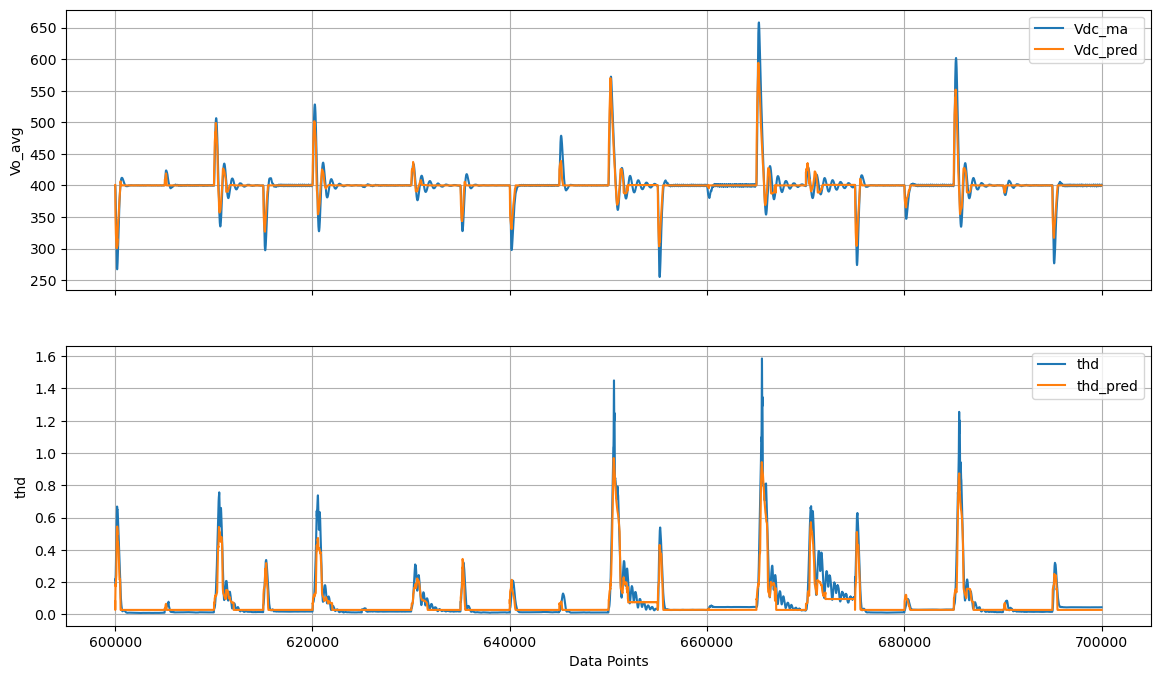}
    \caption{Predicted vs. true transient waveforms ($V_{dc}$ and THD) for a test data subset using $N=350$ uniform random samples.}
    \label{fig:rand_predictions}
\end{figure}

\begin{figure}
    \centering
    \includegraphics[width=1\linewidth]{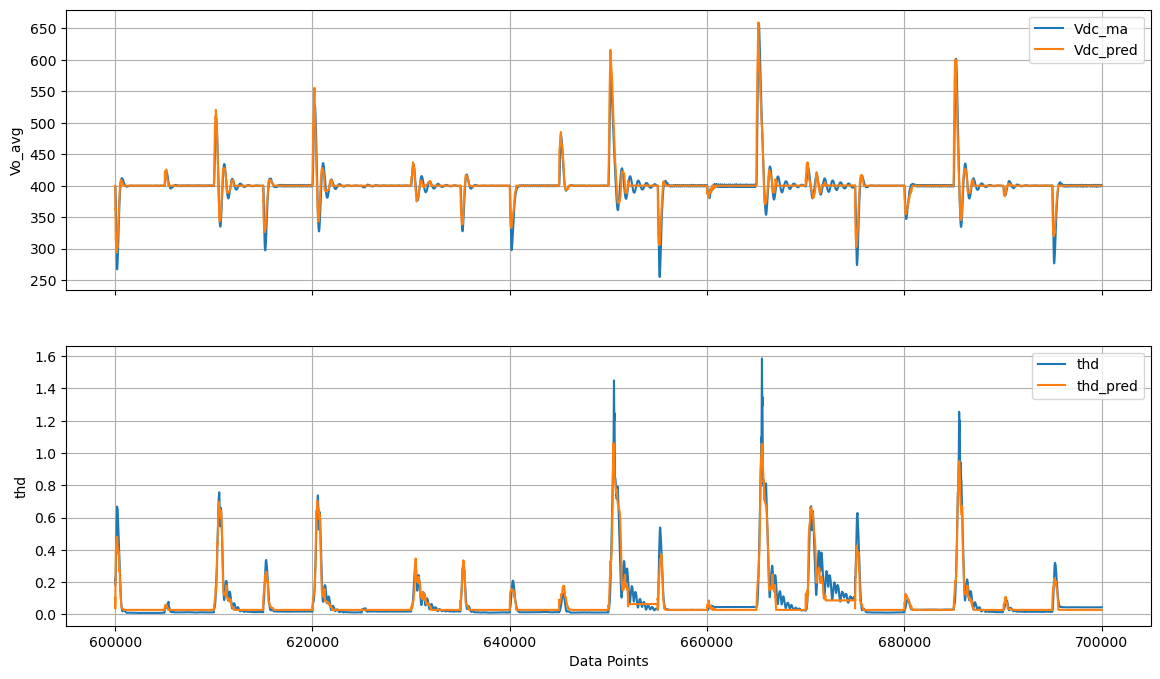}
    \caption{Predicted vs. true transient waveforms ($V_{dc}$ and THD) for a test data subset using $N=350$ Hierarchical GSETS samples.}
    \label{fig:adv_predictions}
\end{figure}

The practical impact of these quantitative disparities is visually apparent in the transient prediction profiles. The surrogate model trained on uniform random data (Fig.~\ref{fig:rand_predictions}) fails to track peak transient deviations accurately and experiences rapid tracking degradation, with the induced transient oscillations dying down prematurely. In multiple instances, the peak dynamic overshoots are omitted entirely. This tracking degradation is absent in the GSETS-trained model (Fig.~\ref{fig:adv_predictions}), despite being trained on an identical data volume of 350 samples. The Hierarchical GSETS framework tracks these transition-induced oscillations with higher structural fidelity as they decay. This performance gap is more pronounced within the output DC-link voltage ($V_{dc}$) transient curves, whereas the tracking improvements within the THD profiles, while present, are less discernible visually.

A primary operational advantage of the GSETS framework is that the generated trajectory sequence is entirely decoupled from the underlying physical surrogate model. Consequently, the trajectory data can be pre-generated independently of the target application and universally reused across different physical systems or simulation configurations. Future work will investigate asymmetrical parameter spacing, allowing individual control axes to maintain independent interval resolutions customized to their unique dynamic sensitivities. Additionally, architectural enhancements will target multi-resolution staging techniques to lower the tensor resolution of the secondary layer, thereby mitigating the space complexity scaling factor while fully preserving the benefits of high-density transition profiling.

\section{Conclusion}
This paper presents Gaussian-Smoothed Efficient Transition Sampling (GSETS), a novel one-shot trajectory generation framework designed to maximize information density and data quality for training data-driven digital twins. By advancing from a global spatial memory heatmap in its base mode to a 4D transition tensor with a dual-kernel spatial spillover mechanism in its hierarchical mode, GSETS successfully steers exploration away from redundant paths. The framework was validated by developing a DNN surrogate for a dual power-sharing T-type rectifier system. Experimental results demonstrate that Hierarchical GSETS achieves high sample efficiency, yielding a 21.0\% reduction in $V_{dc}$ tracking error compared to conventional random sampling under an identical budget. Remarkably, the GSETS framework operating on a sparse allocation of only 350 samples outperforms standard grid and random sampling techniques even when the baselines are granted nearly three times the data volume ($N=1000$).

\section*{Acknowledgment}
This publication was made possible by Qatar National Research Fund under Grant NPRP12C-0814-190012. The statements made herein are solely the responsibility of the authors.

\bibliographystyle{ieeetr}
\bibliography{refs}

\end{document}